\documentclass{article}

 \usepackage[preprint]{neurips_2026}

\usepackage[utf8]{inputenc} 
\usepackage[T1]{fontenc}    
\usepackage{hyperref}       
\usepackage{url}            
\usepackage{booktabs}       
\usepackage{amsfonts}       
\usepackage{nicefrac}       
\usepackage{microtype}      
\usepackage{xcolor}         

 \usepackage{amsmath}

 \usepackage{enumitem}

 \usepackage[utf8]{inputenc}
\usepackage{booktabs} %
\usepackage{tabularx} %
\usepackage{comment}

\usepackage{graphicx}
\usepackage{wrapfig}

\usepackage{etoc}
\usepackage{titletoc}

\title{From Model Scaling to System Scaling: \\ Scaling the Harness in Agentic AI}

\author{%
  Shangding Gu\thanks{This manuscript is under active development, and we welcome any constructive comments and suggestions at \textit{shangding.gu@berkeley.edu}.} \\
  UC Berkeley\\
}

\begin{document}

\maketitle

\vspace{-15pt}

\begin{center}
\textbf{Website:} \url{https://cheetahclaws.github.io}
\end{center}

\vspace{20pt}

\begin{abstract}
This paper studies the next major bottleneck in agentic AI as \emph{system scaling}, not only model scaling: the design of auditable, persistent, modular, and verifiable architectures around foundation models. 
We refer to this shift as \emph{scaling the harness}: treating the structured execution layer around a foundation model as a first-class object of design, evaluation, and optimization. 
Recent progress in large language models (LLMs) has enabled agents that use tools, retrieve information, maintain memory, and execute long-horizon workflows.  Yet evaluation remains largely model-centric, reducing agents to final-task success or benchmark accuracy while treating memory, retrieval, tool use, orchestration, verification, and governance as secondary implementation details. 
This framing is increasingly inadequate: agent performance emerges from the interaction among the foundation model, memory substrate, context constructor, skill-routing layer for tools and subagents, orchestration loop, and verification-and-governance layer. 
Together, these components form the agent harness, the system that translates model capability into long-horizon agent behavior.
We therefore study \emph{scaling the harness} through three core bottlenecks in agentic AI: \emph{context governance}, \emph{trustworthy memory}, and \emph{dynamic skill routing}, together with the orchestration and governance mechanisms that coordinate and constrain them. 
We further outline a research agenda for harness-level benchmarks that operationalize system scaling, going beyond one-shot task success to measure trajectory quality, memory hygiene, context efficiency, communication fidelity, verification cost, and safe evolution over time. Alongside the framework, we develop and release \texttt{CheetahClaws}\footnote{\url{https://github.com/SafeRL-Lab/cheetahclaws}}, a Python-native reference harness, and use it together with Claude Code and OpenClaw as concrete points of comparison that make harness-level design choices explicit. 
Our main claim is that future progress in agentic AI will depend as much on system design as on stronger foundation models.

\end{abstract}

\clearpage
\startcontents[appendix]

\printcontents[appendix]{}{1}{\setcounter{tocdepth}{3}}

\clearpage

\section{Introduction}
\label{sec:intro}

The dominant story of recent AI progress has been \emph{model scaling}: larger models, more data, stronger post-training, and higher benchmark scores~\citep{openai2026gpt54,anthropic2026claudeopus47,google2026gemini31pro}. For agentic AI, this story is now incomplete. Once foundation models are embedded into tools, terminals, browsers, repositories, memory stores, and external services, their behavior is no longer determined by the model alone. It is determined by a \emph{system}: how context is constructed, how memory is retrieved, how tools are invoked, how subagents are routed, how actions are verified, and how failures are audited.

Our key claim is therefore that \textbf{agentic AI should be studied and evaluated as a system-scaling problem, not merely as a model-scaling problem.} By \emph{model scaling}, we refer to improvements in the standalone foundation model, including model size, training data, post-training, and raw reasoning capability. By \emph{system scaling}, we refer to improvements in the surrounding architecture, including memory, context construction, skill routing across tools and subagents, orchestration, and verification-and-governance, and how these components adapt over time. Equivalently, this is a problem of \emph{scaling the harness}: improving the structured execution layer around the foundation model, so that these system components work reliably over long horizons. Our claim is not that model scaling no longer matters; rather, once models reach a sufficient capability threshold, many additional gains in long-horizon agent performance increasingly depend on how the system around the model is designed.

Modern agentic systems already illustrate what scaling the harness looks like in practice. Production harnesses such as Claude Code~\citep{claude-code-2025} and OpenClaw~\citep{openclaw2026} couple foundation models to tools, subagents, and persistent project memory (detailed in §\ref{sec:harnesses}); research-side harnesses such as SWE-agent further show that careful tool-schema design alone can improve benchmark accuracy substantially even with a fixed backbone model~\citep{yang2024sweagent}. These systems show that practical agent capability does not arise from next-token prediction alone, but from the interaction between the foundation model and the harness that surrounds it. The relevant object of study is therefore not simply a model plus prompt, but a structured execution system, a view increasingly reflected in recent work on code-centered agent harnesses~\citep{ning2026code}.

This perspective is highlighted by recent empirical findings. A field-level analysis of agent benchmarks finds that many results do not separate capability from costs, prompting strategy, and demonstrations, and become non-Pareto-optimal once these factors are controlled~\citep{kapoor2024aiagents}. Consistent with this, redesigning the agent--computer interface alone, while holding the underlying model fixed, can substantially improve SWE-bench accuracy~\citep{yang2024sweagent}. Thus, what is often reported as a model score is in fact a model-plus-harness score. Context length is another example: larger context windows do not guarantee effective information access, because attention dilutes over long inputs~\citep{gu2026long}, and models often prefer evidence at the start or end of the context rather than in the middle~\citep{liu2024lost}. Multi-agent systems show a similar pattern: they can outperform single agents on breadth-first tasks but introduce coordination failures that single-agent metrics miss~\citep{anthropic_multiagent, cemri2025multiagent}; we return to this in §\ref{sec:longitudinal}. Realistic agent benchmarks such as GAIA~\citep{mialon2024gaia}, $\tau$-bench~\citep{yao2024taubench}, and Terminal-Bench~\citep{merrill2026terminal} further show that frontier models struggle when evaluation moves from one-shot prompting to multi-step interaction with tools, environments, and users. In particular, $\tau$-bench shows that agents that look strong under single-shot pass rates can collapse under $\text{pass}\char`\^ k$, the probability of succeeding on $k$ independent rollouts. This exposes a reliability gap that endpoint accuracy hides.

These findings suggest that we need to rethink several parts of the agent system. Prompt engineering~\citep{white2023prompt} remains useful for local control, but long-horizon performance increasingly depends on reusable skills, persistent memory, disciplined context construction, and verification-aware execution. The key issue is not only context size, but \emph{context governance}: what should be retrieved, compressed, ordered, refreshed, trusted, and kept active at each step. Memory is not merely a storage layer; the harder problem is memory \emph{quality}, including what to store, what to discard, how to retrieve the right information at the right time, and how to avoid staleness, drift, contamination \citep{altawaha2026rememberingmoreriskingmore}, and over-generalization. Multi-agent systems are not automatically collaborative; reliable collaboration requires explicit communication protocols and uncertainty sharing~\citep{guo2026llms}, which we expand on in §\ref{sec:longitudinal}. Finally, the field still lacks a mature framework for \emph{agent evolution} over time, including how agents should update skills, refine memory, communicate across roles, and remain auditable as they adapt.

This paper makes three main contributions:
\begin{itemize}[leftmargin=1.5em]
    \item \textbf{System-scaling framing.} We develop a systems-centered framing of agentic AI in which progress depends on \emph{scaling the harness}, not only scaling the model. Our main claim is that the next bottleneck in agentic AI is not only how powerful the model is, but how well the surrounding system manages memory, context, skill routing across tools and subagents, orchestration, verification and governance, and adaptation over time.

    \item \textbf{Harness-level framework.} We propose a framework that separates base-model reasoning from system factors including memory, context construction, skill routing, orchestration, and verification-and-governance. This framework treats the agent harness as a first-class object of design and analysis.

    \item \textbf{Evaluation agenda and reference harness.} We outline an evaluation agenda for agentic systems, highlighting that future benchmarks should measure process-level and longitudinal properties such as trajectory quality, memory hygiene, context efficiency, verification cost, safe evolution, and robustness under repeated use. To make the discussion concrete, we develop \texttt{CheetahClaws}, a Python-native reference harness, and compare it against Claude Code and OpenClaw, treating their harness-level design choices as instances of the system-scaling variables identified by our framework.
\end{itemize}

\section{Related Work}
\label{sec:related}

\paragraph{Agentic coding systems and harness engineering.}
Modern coding agents follow a line of work on tool-using language models, beginning with interleaved reasoning--and--acting policies such as ReAct~\citep{yao2023react}, self-taught tool invocation~\citep{schick2023toolformer}, and verbal self-correction loops~\citep{shinn2023reflexion}. Production systems such as Claude Code~\citep{claude-code-2025, anthropic_autonomy} and Codex-style ``harness engineering''~\citep{Harness_engineering} package these primitives into programmable agent runtimes with tools, subagents, hooks, and persistent project memory. A parallel research line targets software engineering specifically, including SWE-agent's agent--computer interface, which shows that carefully designed tool schemas can by themselves move benchmark accuracy substantially even with a fixed backbone model~\citep{yang2024sweagent}. Most of this work, however, reports results at the level of individual model variants; comparatively little attention has been paid to the \emph{harness itself} as a controllable, reproducible object of study, which is the vantage we adopt throughout this paper.

\paragraph{Context, memory, and retrieval.}
Retrieval-augmented generation~\citep{lewis2020rag} showed that augmenting parametric language models with external non-parametric memory can substantially improve knowledge-intensive generation and question answering. And following work studies memory as a system component, including MemGPT's hierarchical memory management~\citep{packer2023memgpt} and Voyager's growing skill library for open-ended exploration~\citep{wang2023voyager}. At the same time, recent analyses show that longer context windows come with their own failure modes such as privacy drift~\citep{gu2026long}, and that agents still need calibrated uncertainty to decide when to retrieve at all~\citep{guo2026llms}. These results motivate our treatment of context, memory, and retrieval as a \emph{context-governance} problem rather than as independent capabilities.

\paragraph{Skills and multi-agent coordination.}
Reusable skills have emerged as a way to offload recurring behavior from prompts into durable, callable components~\citep{openai_skills_blog, openai_skills_api, wang2023voyager}, extending earlier work on chain-of-thought prompting~\citep{wei2022chain} and prompt-pattern catalogs~\citep{white2023prompt}. In parallel, multi-agent frameworks such as AutoGen~\citep{wu2023autogen}, MetaGPT~\citep{hong2024metagpt}, and CAMEL~\citep{li2023camel} formalize agent-to-agent communication, while Anthropic reports substantial gains from orchestrator-plus-subagent configurations on breadth-first research tasks~\citep{anthropic_multiagent}. Complementary work studies how population diversity~\citep{yang2026understanding, ye2025x} and negotiation-style frameworks~\citep{liu2026agenticpay} shape collective behavior, and how such agents compose into a broader ``agentic web''~\citep{yang2025agentic}. Our framing treats skills and delegation jointly as the \emph{skill} lever and emphasizes that skill routing under heterogeneous subagents, rather than the existence of skills or subagents, is the next open systems bottleneck.

\paragraph{Benchmarks, governance, and agent evolution.}
A growing line of work evaluates agents as systems through executable, multi-step benchmarks~\citep{jimenez2024swebench, liu2023agentbench, zhou2024webarena, merrill2026terminal}, alongside broader surveys of LLM-based agents~\citep{xi2023rise} and catalogues of agentic safety threats~\citep{owasp2025agentic}; yet single-episode success still dominates the reported metrics, leaving memory quality, context efficiency, communication fidelity, and safe evolution under repeated use largely unmeasured (we return to these in §\ref{sec:evaluation}). Compared to these lines of work, our contribution is to reframe prior developments through a system-scaling perspective and to make its engineering content concrete through a comparative analysis of Claude Code, OpenClaw, and our Python-native reference harness CheetahClaws.

\section{System Scaling: A Framework for Agentic AI}
\label{sec:framework}

Throughout this paper, we use \emph{harness} to refer to the structured system layer surrounding a foundation model: the tool interface, control loop, context constructor, memory store, skill-routing mechanism, and verification-and-governance layer that together mediate between user intent, model outputs, and the external environment. The harness is what model scaling does not include and what system scaling targets.

\begin{figure}[ht]
\centering
\includegraphics[width=1.0\linewidth]{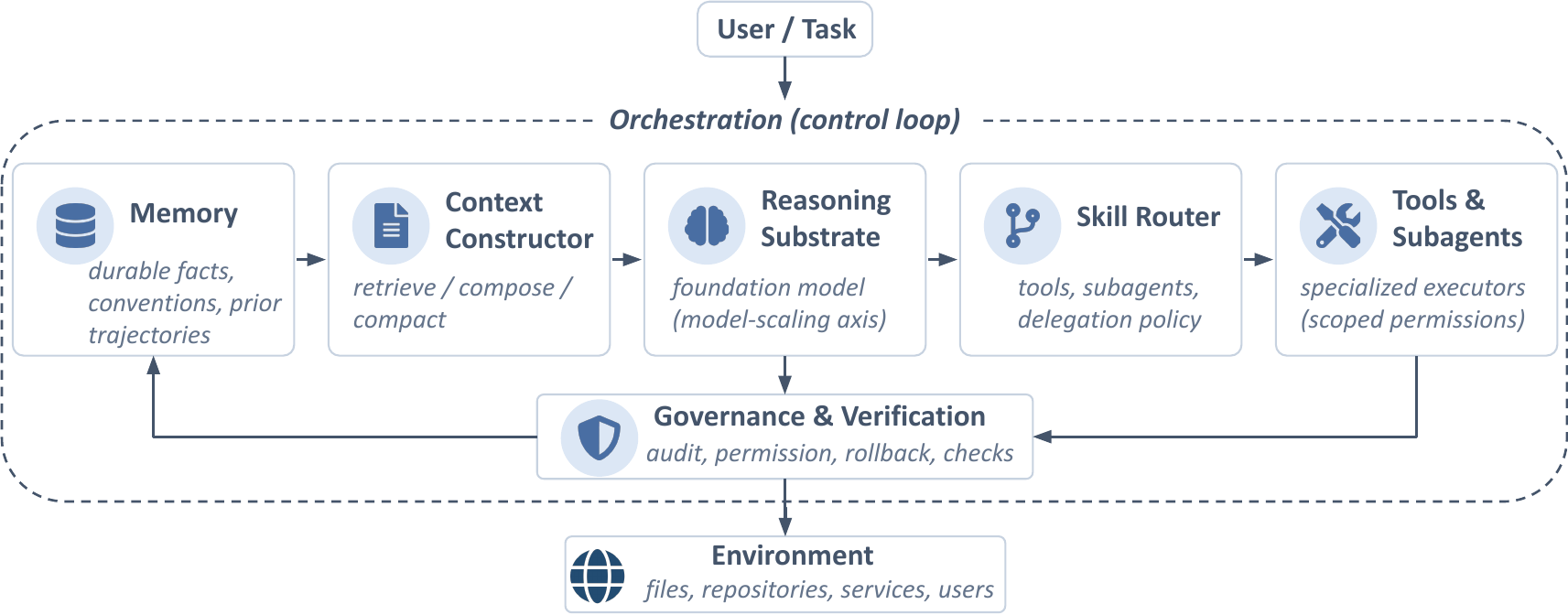}
\caption{A six-component view of an agentic system: $\mathcal{P}_H = \Phi(\mathcal{R}, \mathcal{M}, \mathcal{C}, \mathcal{S}, \mathcal{O}, \mathcal{G})$. The orchestration layer ($\mathcal{O}$) wraps a control loop in which the context constructor ($\mathcal{C}$) draws from durable memory ($\mathcal{M}$) and the current task to assemble inputs for the reasoning substrate ($\mathcal{R}$, i.e.\ the foundation model). The skill router ($\mathcal{S}$) dispatches tools or subagents; their effects on the environment, together with the model's intermediate steps, are gated through verification and governance ($\mathcal{G}$) before they become permitted actions or verified memory write-backs. Model scaling improves $\mathcal{R}$; system scaling improves $\mathcal{M}, \mathcal{C}, \mathcal{S}, \mathcal{O}$, and $\mathcal{G}$.}
\label{fig:system-architecture}
\end{figure}

We use \emph{system scaling} to denote improvements in this harness that determine how information, computation, authority, and verification are allocated over time, and refer to this engineering agenda as scaling the harness. Under this view, an agent is not simply a model with a prompt, but a system composed of six interacting components: a reasoning substrate ($\mathcal{R}$), a memory store ($\mathcal{M}$), a context constructor ($\mathcal{C}$), a skill-routing layer ($\mathcal{S}$, which dispatches tools and subagents), an orchestration loop ($\mathcal{O}$), and a verification and governance layer ($\mathcal{G}$). Let performance over a horizon $H$ be
\begin{equation}
\label{eq:system-scaling}
    \mathcal{P}_H = \Phi(\mathcal{R}, \mathcal{M}, \mathcal{C}, \mathcal{S}, \mathcal{O}, \mathcal{G}),
\end{equation}
where $\mathcal{R}$ denotes base reasoning quality, $\mathcal{M}$ memory quality, $\mathcal{C}$ context-construction quality, $\mathcal{S}$ skill selection and composition quality, $\mathcal{O}$ orchestration quality, and $\mathcal{G}$ governance quality. Model scaling primarily improves $\mathcal{R}$; system scaling improves $\mathcal{M}, \mathcal{C}, \mathcal{S}, \mathcal{O}$, and $\mathcal{G}$. The main claim of this paper is that, once models reach a sufficient capability level, long-horizon agent performance may be limited not only by $\mathcal{R}$ itself, but also by the surrounding system factors. A useful further factorization is
\begin{align}
\label{eq:system-factor}
    \mathcal{M} &= (\text{precision},\, \text{durability},\, \text{retrievability},\, \text{verifiability}), \\
    \mathcal{C} &= (\text{relevance},\, \text{compactness},\, \text{traceability},\, \text{refresh policy}).
\end{align}
Each factor names a system-level lever, not a hidden engineering detail. Figure~\ref{fig:system-architecture} sketches how these components interact: the orchestration loop $\mathcal{O}$ wraps a flow in which $\mathcal{C}$ draws from $\mathcal{M}$ to assemble inputs for $\mathcal{R}$, $\mathcal{S}$ dispatches tools and subagents, and $\mathcal{G}$ gates both intermediate reasoning and external action before any verified result is written back to memory.

\paragraph{Status of the decomposition.} Equation~\ref{eq:system-scaling} is a conceptual organization rather than a quantitative model: $\Phi$ has no closed form, the factors are not strictly orthogonal, and we do not claim they jointly determine $\mathcal{P}_H$ as a measurable equation. What we do claim is that each factor names a distinct point of \emph{intervention}, a place where engineering or research effort changes long-horizon behavior, and that existing discussions frequently fail to distinguish between them. We choose these six axes because each one can be changed, turned off, or measured on its own, while keeping the same foundation model. For instance, run $\mathcal{O}$ in a one-shot loop, or turn $\mathcal{G}$ off, and the same $\mathcal{R}, \mathcal{C}, \mathcal{S}$ start to act like noticeably different agents. Among the six, $\mathcal{R}$ and $\mathcal{C}$ are the hardest to separate (a stronger reasoning substrate can compensate for noisier context, and vice versa), while $\mathcal{M}$ and $\mathcal{G}$ are the easiest to isolate, since they govern writes and audit trails that exist independently of any single inference step.

\subsection{Agent Harnesses as System Infrastructure}
\label{sec:harnesses}

Modern agent harnesses such as OpenClaw~\citep{openclaw2026} and Claude Code~\citep{claude-code-2025} are better understood as \emph{systems infrastructures} rather than simple model interfaces: their behavior depends not only on the underlying language model, but on the surrounding tool interface, execution loop, context constructor, memory substrate, and orchestration policy. Claude Code in particular benefits from substantial harness engineering~\citep{Harness_engineering}: it bundles tools for codebase navigation, file editing, and command execution; dispatches specialized subagents with their own context windows, prompts, and permissions; and adopts a hybrid context strategy that loads persistent project guidance up front while retrieving information just in time through \texttt{glob}/\texttt{grep}-style tools.\footnote{See documentation at \url{https://code.claude.com/docs/en/overview} (overview), \url{https://code.claude.com/docs/en/sub-agents} (subagents), and \url{https://platform.claude.com/docs/en/agent-sdk/python} (SDK).} What distinguishes modern agentic coding systems from classic code assistants is therefore not stronger token-level generation alone, but the presence of an execution harness that supports tool use, iterative verification, and task decomposition.

These details matter because they shift attention from model capability alone to the system conditions under which that capability is expressed.  The relevant unit of analysis is not simply an isolated foundation model conditioned on a prompt, but the interaction among the six components introduced in Equation~\ref{eq:system-scaling}: $(\mathcal{R}, \mathcal{M}, \mathcal{C}, \mathcal{S}, \mathcal{O}, \mathcal{G})$. These are not minor implementation details. They determine what information is available at decision time, how external actions are executed and verified, and how progress accumulates across turns. As a result, they increasingly govern task-level performance in long-horizon settings. Once these components are treated as first-class objects, the key research question shifts from \textbf{``\textit{How do we prompt the model better?}''} to \textbf{``\textit{How do we allocate computation across memory, retrieval, tools, and subagents over time?}''}

\begin{table}[htb]
\centering
\caption{Illustrative harness design patterns. The point is not to rank systems, but to show that comparable agent primitives can be governed differently under different deployment priorities.}
\label{tab:harness-comparison}
\small
\setlength{\tabcolsep}{4pt}
\renewcommand{\arraystretch}{1.15}
\begin{tabularx}{\linewidth}{@{}p{2.5cm}XXX@{}}
  \toprule
                    & \textbf{Claude Code} & \textbf{OpenClaw} & \textbf{CheetahClaws} \\
  \midrule
  Implementation    & TypeScript           & TypeScript        & Python \\
  Primary setting   & Vendor coding agent  & Personal assistant & Research reference \\
  Primary user
  interaction       & Terminal CLI / IDE   & Messaging app (Discord, Slack, iMessage, \ldots) & Terminal CLI \\
  Context governance & User, project, session & User, channel-peer, session & User, project, session \\
  Memory            & Persistent text memory, auto-extraction & Conversation history, vector retrieval & Structured entries with confidence, recency \\
  Source availability      & Closed-source & Open-source  & Open-source \\
  \bottomrule
  \end{tabularx}
\end{table}

A natural skeptical view holds that, once the foundation model is held fixed, most harnesses collapse to the same tool loop, with only superficial differences between them. We show instead that the similar core systems problems, context governance, memory trust, skill routing, and auditability, admit substantially different solutions depending on deployment priorities. Table~\ref{tab:harness-comparison} sketches three illustrative design points built around comparable frontier-model capabilities: \textbf{Claude~Code} (v2.1.88), a production-grade vendor harness; \textbf{OpenClaw} (v2026.4.6), a community TypeScript harness for multi-channel personal assistance; and \textbf{CheetahClaws} (v3.05.79), a Python-native reference harness used here as an open illustrative design point. Two observations follow. First, the three systems reflect a shared systems-decomposition principle: each addresses context governance, memory management, and skill routing, even though these levers are realized through different design choices. This convergence suggests that they are intrinsic design problems for agentic AI systems, rather than incidental features of any particular implementation. Second, their main differences are driven less by the foundation model than by deployment priorities: vendor-scale systems prioritize reliable use, personal-assistant systems prioritize a gateway for multi-channel management, and research-oriented harnesses prioritize transparency and reproducibility. The remainder of the paper makes these levers concrete: §\ref{sec:bottlenecks} expands the three bottlenecks of context, memory, and skill, and §\ref{sec:evaluation} discusses how to evaluate and govern their evolution.

All three systems consolidate session content into persistent memory through subagent extraction, background daemons, or dedicated consolidation routines. What differs is the representation of trust: CheetahClaws stores per-entry confidence and recency as first-class fields, used directly in retrieval ranking and conflict resolution. The other two derive trust implicitly from access patterns. In this sense CheetahClaws operationalizes the trust axes of §\ref{sec:trust-memory} more directly.

\subsection{Prompt, Skill, and Memory as Temporal Layers}
\label{sec:psm}

We interpret prompt, skill, and memory as three primary \emph{temporal} axes of system scaling in agentic AI. This view is complementary to Equation~\ref{eq:system-scaling}: skill corresponds to $\mathcal{S}$ and memory to $\mathcal{M}$, while prompt sits inside each per-turn output of the context constructor $\mathcal{C}$; the orchestration $\mathcal{O}$, verification and governance $\mathcal{G}$ layers determine how the three are sequenced and verified over time. As shown in Table~\ref{tab:psm}, they operate at different temporal scales and support different forms of adaptation.

\paragraph{Prompt.} Prompt is the short-horizon control interface. It specifies the immediate role, constraints, and objective. Prompting is flexible and cheap, but also brittle: it does not by itself create persistence, transfer, or reliable long-horizon structure.

\paragraph{Skill.} A skill is a reusable execution pattern. In practice, a skill may appear as a workflow template, a tool-use routine, a specialized subagent, or a versioned bundle of instructions and scripts. OpenAI's recent discussions of skills for coding agents make this direction explicit: durable procedures are separated from one-off prompts and packaged as reusable components attached to the execution environment~\citep{openai_skills_blog, openai_skills_api}. Skills make behavior more reusable, but introduce a routing problem: the agent must decide which skill to invoke, when to switch skills, and how to compose multiple skills in one trajectory.

\paragraph{Memory.} Memory is the longitudinal layer. It stores what should persist across turns or sessions: project conventions, user preferences, stable facts about the environment, prior failures, and distilled structure from earlier work. Memory is essential for repeated tasks, but it can fail along three trust axes elaborated in Section~\ref{sec:trust-memory}: drift (loss of durability), over-generalization (loss of precision), and pollution (loss of verifiability).

These three levers are complementary rather than interchangeable. Prompt controls \emph{\textbf{what to do now}}; skill controls \emph{\textbf{how to do this class of things}}; memory controls \emph{\textbf{what should survive over time}}. A robust agent is therefore not merely well prompted. It is well prompted \emph{and} appropriately skilled \emph{and} selectively grounded in durable memory.

\begin{table}[t]
\centering
\caption{Prompt, skill, and memory as three core axes of system scaling in long-horizon agents.}
\label{tab:psm}
\resizebox{0.99\columnwidth}{!}{%
\begin{tabular}{p{2cm}p{2.7cm}p{4.2cm}p{4.3cm}}
\toprule
\textbf{Lever} & \textbf{Timescale} & \textbf{Primary role} & \textbf{Typical failure mode} \\
\midrule
Prompt & Local & Specify current goal, constraints, and style & Brittle over long horizons; poor transfer \\
Skill & Task-level & Reusable procedure or workflow pattern & Wrong routing or poor composition \\
Memory & Longitudinal & Preserve durable facts and prior experience & Drift, over-generalization, pollution (durability / precision / verifiability) \\
\bottomrule
\end{tabular}
}
\end{table}

\section{Three Bottlenecks in System Scaling}
\label{sec:bottlenecks}

We now expand three system factors from Equation~\ref{eq:system-scaling} where model scaling alone has been least sufficient: context construction $\mathcal{C}$, memory $\mathcal{M}$, and skill routing $\mathcal{S}$, tightly coupled to verification and governance $\mathcal{G}$. Each subsection names four subaxes of its component, the dominant failure mode, and the system move that addresses it; Table~\ref{tab:bottleneck-template} summarizes the three.

\begin{table}[t]
\centering
\small
\caption{Three bottlenecks of system scaling. Each subsection names four subaxes of one component, a characteristic failure mode, and the system move that addresses it.}
\label{tab:bottleneck-template}
\setlength{\tabcolsep}{6pt}
\renewcommand{\arraystretch}{1.2}
\resizebox{0.99\columnwidth}{!}{%
\begin{tabular}{@{}lp{4cm}lp{3.8cm}@{}}
\toprule
\textbf{Component} & \textbf{Subaxes} & \textbf{Failure mode} & \textbf{System move} \\
\midrule
$\mathcal{C}$ governance (§\ref{sec:context-governance}) & relevance, compactness, traceability, refresh & exposure without access & assembly as a policy; persistent priors plus just-in-time refresh \\
$\mathcal{M}$ trust (§\ref{sec:trust-memory}) & precision, durability, retrievability, verifiability & stale-but-confident & trust re-established at retrieval; periodic verification against the environment \\
$\mathcal{S}$ routing (§\ref{sec:skill-routing}) & specificity, selectivity, composability, verifiability & confident-but-unchecked & adaptive routing coupled with explicit post-condition checks \\
\bottomrule
\end{tabular}
}
\end{table}

\subsection{Context Governance}
\label{sec:context-governance}

\textbf{The hard problem of context is not capacity, but \emph{governance}.} From the four axes of $\mathcal{C}$ in Equation~\ref{eq:system-factor}, an effective context assembly is jointly \emph{relevant} to the current task, \emph{compact} (no more than the minimum sufficient set), \emph{traceable} to its sources, and refreshed against a moving environment. Larger context windows expand capacity, but they do not guarantee relevance, compactness, traceability, or freshness.

The threat we are guarding against is \emph{exposure without access}: as context grows, the model sees more tokens but does not necessarily attend to the right ones. Relevant evidence competes with low-value padding (signal dilution~\citep{gu2026long}), task-relevant structure is buried in unorganized text, and token salience is driven by local statistics rather than decision importance. Long context does not indicate good context; tokens added without governance often degrade performance rather than improve it.

The system move is to treat each turn's context as the output of a selection policy, not a fixed buffer. The policy should weight semantic relevance, penalize verbosity against a token budget, prefer recently validated content, and record provenance so failures can be attributed at audit time. The right systems question is therefore not how many tokens the model can hold, but how the system constructs the \emph{minimum sufficient context} for the current subproblem.

\subsection{Trustworthy Memory}
\label{sec:trust-memory}

\textbf{The hard problem of agent memory is not storage, but \emph{trust}.} Matching three of the four axes of $\mathcal{M}$ in Equation~\ref{eq:system-factor}, a memory item earns trust when it is \emph{precise} within a defined scope, remains \emph{durable} (its target has not silently drifted), and is \emph{verifiable} against the current environment. The fourth axis, retrievability, controls whether that trust can be used at acceptable cost. It is a precondition for using trust, not a source of trust.

The threat we are guarding against is \emph{stale-but-confident}. A note that was correct at one point, say ``the data loader is defined in \texttt{utils/loader.py}'', can become flatly wrong after a refactor without any change to its wording. Semantic search and reuse statistics still rank it highly, but its target has drifted, and acting on it is now destructive (calling a deleted symbol, or reintroducing a fixed regression). The failure mode is asymmetric: stale memory rarely prevents retrieval, but regularly leads the agent to act confidently on invalidated assumptions.

The system move is to make trust a runtime decision, not a property of the stored item. Retrieval should weight a staleness penalty (against the time of last verification) and a confidence-gated risk term alongside any relevance score, and should treat the retrieved content as a hypothesis until re-checked against the live environment. Claude Code realizes this coupling through a hybrid design: 
\texttt{CLAUDE.md} carries persistent project context, while 
built-in primitives (\texttt{glob}, \texttt{grep}, file reads) 
provide just-in-time access to the live repository, so the 
agent can re-verify environment-dependent facts on demand 
instead of trusting a static index~\citep{anthropic_context_engineering,context_engineering_he2026, claude_code_memory_docs}. Durable memory without periodic verification accumulates undetected drift; environment-only search without distilled priors discards every prior verification. Trustworthy memory keeps both: it retains accumulated verification while bounding drift.

\subsection{Dynamic Skill Routing and Verification}
\label{sec:skill-routing}

\textbf{The hard problem of skill is not having skills, but routing and checking them.} Extending the factorization in Equation~\ref{eq:system-factor} to $\mathcal{S}$, effective skill use requires four conditions: each skill is \emph{specific} about its capability scope, the routing policy is \emph{selective} in invoking the right skill, the skill set is \emph{composable} (one skill's post-conditions feed the next), and every skill output is \emph{verifiable} against an explicit check.

The threat we are guarding against is \emph{confident-but-unchecked}: a specialized subagent can return plausible output that no downstream layer validates. As specialized skills multiply, the failure mode shifts from a missing capability to a present-but-unverified one. This is the symmetric form of stale-but-confident memory in §\ref{sec:trust-memory}: both let the agent act on a claim whose truth condition was never re-established.

The system move is to treat routing as a learned policy, not a fixed rule set, coupled with verification at every step. Dynamic skill routing is the analogue of scheduling in operating systems: raw skill capacity exists, but useful work depends on allocating it to the right specialized pathway at the right time. The open research direction is to make this allocation adaptive through online estimates of subtask type, confidence-aware escalation, mixture-style composition, and policies optimized for verified rather than fluent intermediate outputs; and to make post-condition checking a first-class component of each skill specification. In the notation of Equation~\ref{eq:system-scaling}, $\mathcal{S}$ and $\mathcal{G}$ are therefore not independent: scaling skill quality without scaling verification produces faster but less reliable progress.

\section{Toward System-Level Evaluation and Agent Evolution}
\label{sec:evaluation}

\subsection{From Outcome Metrics to Process Metrics}
\label{sec:outcome-to-process}

Benchmarks for agentic AI have improved rapidly, and the current generation already gets several important things right. SWE-bench~\citep{jimenez2024swebench} demonstrated that executable, repository-level tasks can be evaluated through their own test suites, anchoring agent evaluation in real codebases; AgentBench~\citep{liu2023agentbench} pushed evaluation across diverse interactive environments rather than a single domain; WebArena~\citep{zhou2024webarena} did the similar for browser-based agents under realistic distributions; and Terminal-Bench~\citep{merrill2026terminal} most recently introduces hard, environment-grounded terminal tasks with per-task verification. These benchmarks have collectively moved evaluation away from static next-token prediction and toward multi-step execution against real artifacts, and our claim is not that they are wrong.

However, they remain \emph{insufficient} for evaluating system-scaled agents. As noted in §\ref{sec:intro}, single-score reporting may mix model capability with harness design, making it difficult to tell whether gains come from a stronger model or a better system around it. The problem becomes more serious in long-horizon and multi-agent settings, where small system choices, including which files to inspect first, which facts to retain, when to run tests, and how to recover from failed actions, can accumulate over time and shape the final outcome (see §\ref{sec:longitudinal} for the multi-agent case). Endpoint metrics also fail to capture cost and risk. For example, two agents may both solve a task, yet differ substantially in tokens, tool calls, retries, failed edits, human interventions, and auditability. These process-level differences determine latency, monetary cost, user trust, reproducibility, and deployment safety.

A stronger protocol should therefore report \emph{outcome metrics} (whether the task was solved) jointly with \emph{process metrics} (how much context and computation were used, how the trajectory unfolded, what was retrieved and verified, and how risk was incurred), so that the system factors in Equation~\ref{eq:system-scaling} can be measured rather than hidden. The aim is to extend the evaluation surface that SWE-bench, AgentBench, WebArena, and Terminal-Bench have opened up, not to replace it.

\subsection{From Single Episodes to Longitudinal Evaluation}
\label{sec:longitudinal}

Multi-agent systems illustrate why agent evaluation needs to move beyond single-episode success. Anthropic reports that a multi-agent system (Claude Opus 4 lead agent with Claude Sonnet 4 subagents) outperformed single-agent Claude Opus 4 by 90.2\% on their internal research evaluation, with multi-agent architectures being especially effective for breadth-first tasks that explore several independent directions in parallel. In their BrowseComp-based analysis, token usage alone accounted for 80\% of the performance variance, and adding tool-call count and model choice raised the explained variance to 95\%~\citep{anthropic_multiagent}. These results suggest that multi-agent architectures can provide useful additional compute by exploiting parallel context windows and task decomposition, while also showing that agent performance is strongly shaped by how the system allocates computation and tool use.

However, this does not imply that collaboration emerges automatically. Decomposition is easier than collaboration, and recent failure analyses show that current multi-agent systems often fail because of system-design issues, inter-agent misalignment, and inadequate task verification, rather than merely because of underlying model limitations~\citep{cemri2025multiagent}. True collaboration requires shared state, uncertainty communication, contradiction detection, task de-duplication, and conflict resolution. The real open problem is therefore not whether multiple agents can be wired together, but whether the communication protocol between them can be made reliable enough for long-horizon work; handoffs, summaries, requests for clarification, and uncertainty reports should be treated as optimized objects rather than ad hoc prompt fragments.

This points to a more general gap: most current benchmarks reset the agent between tasks, but real agents accumulate state across sessions, conversations, and projects. They store conventions, preferences, prior failures, and reusable procedures, and the same accumulation that enables improvement can also produce failure modes such as contamination, staleness, over-generalization, and privacy leakage. A one-shot evaluation cannot reveal whether an agent's memory becomes more useful, more noisy, or more dangerous over repeated use.

We list these dimensions in Table~\ref{tab:benchmark-dimensions}. The next generation of agent benchmarks should additionally measure repeated-use properties such as memory retrieval precision and memory hygiene, minimal-context efficiency, communication fidelity across subagents, drift across long trajectories or sessions, verification-aware recovery after stale memory or wrong routing, and safety under tool access and autonomous execution. Current evaluation often measures whether an agent can finish \emph{a} task, but not whether it can finish similar tasks repeatedly while improving, staying grounded, and avoiding silent degradation. Agent quality should therefore be evaluated as a \emph{longitudinal systems property} rather than a one-shot completion score.

\begin{wraptable}{r}{0.52\columnwidth}
\vspace{-1.2em}
\centering
\small
\caption{Benchmark dimensions for system-scaling evaluation.}
\label{tab:benchmark-dimensions}
\renewcommand{\arraystretch}{1.05}
\setlength{\tabcolsep}{3pt}
\resizebox{0.45\columnwidth}{!}{%
\begin{tabular}{lcc}
\toprule
\textbf{Dimension} & \textbf{Typical Benchmarks} & \textbf{Needed} \\
\midrule
One-shot completion        & Common    & Yes \\
Memory retrieval precision & Rare   & Yes \\
Memory hygiene             & Rare   & Yes \\
Minimal-context efficiency & Rare   & Yes \\
Communication fidelity     & Rare   & Yes \\
Long session/trajectory drift   & Rare   & Yes \\
Verification-aware recovery& Partial     & Yes \\
Safety under tool access   & Partial & Yes \\
\bottomrule
\end{tabular}
}
\vspace{-1.2em}
\end{wraptable}

\subsection{Standards for Safe Agent Evolution}
\label{sec:safe-evolution}

A mature agent should not only act, but evolve. Yet the field lacks a standard for what persistent adaptation should mean. What should be allowed to change over time, memory only, or also routing policies, skills, and collaboration protocols? What should be fixed for auditability? What counts as safe improvement versus dangerous drift? The questions are not abstract: persistent behaviors can survive subsequent training in ways that are hard to detect from outputs alone~\citep{hubinger2024sleeper}; optimization against imperfect proxy objectives can induce characterizable reward-hacking failure modes~\citep{skalse2022defining}; and the OWASP catalogue of agentic threats lists memory poisoning, identity spoofing, tool misuse, and goal manipulation as exploitable failure surfaces~\citep{owasp2025agentic}. These failures become especially salient as agents evolve. These are exactly the failure modes a maturity standard for agent evolution would have to make visible, measurable, and bounded.

We therefore propose that future agent systems need an explicit \emph{agent evolution standard} built around four questions:
\begin{enumerate}[leftmargin=1.5em]
    \item \textbf{What persists?} Memory, skills, preferences, and guardrails should be distinguished rather than merged into one undifferentiated state, so that updates to one component do not silently rewrite another.
    \item \textbf{What updates?} Update policies should distinguish components that may adapt online from those requiring review, replay, or stronger verification, especially when changes interact with tool permissions or governance boundaries.    
    \item  \textbf{What is measured?} Longitudinal improvement should be assessed together with regression, drift, and the recurrence of earlier failures, rather than inferred from a single rolling success rate. Evaluation should also account for reward-hacking failure modes that arise when agents optimize imperfect proxy objectives~\citep{skalse2022defining}.   
    \item \textbf{What is auditable?} Memory writes, routing changes, tool permissions, and collaboration failures should leave inspectable traces \citep{owasp2025agentic}. Behavioral evaluation alone is insufficient for persistent risks of the kind documented in \citep{hubinger2024sleeper}, where backdoored behaviors survive SFT, RL, and adversarial training.
\end{enumerate}

Without such standards, many so-called learning agents risk becoming opaque accumulations of prompts, notes, and heuristics rather than reliable adaptive systems.

\section{Discussion: Alternative Views and Limitations}
\label{sec:counterarguments}

The system-scaling claim is incomplete without an honest engagement with the views it stands against. We discuss three objections.

\paragraph{Objection 1: Stronger models will eventually solve system problems.}
One objection is that system scaling is a temporary concern: as foundation models become stronger, they may learn to manage context, memory, tools, and verification internally, without an explicit harness. We agree that model scaling will continue to improve agent behavior. However, many failures in deployed agents are not failures of next-token prediction alone. Stale memory, over-broad tool permissions, missing provenance, unverified retrieval, and unsafe action execution are system failures. A stronger model may reduce their frequency, but it does not remove the need for explicit mechanisms that govern what information is exposed, which actions are authorized, and how failures are traced. Whatever the model's capability, an agent that can act on the world requires a system around it that decides what actions are permitted and how they are verified.

\paragraph{Objection 2: End-to-end training will replace modular systems.}
A second view is that future agents should be trained end-to-end, making explicit system components unnecessary. End-to-end training may improve coordination across components, but deployed agents still require modular boundaries. They operate over private files, credentials, tools, repositories, browsers, and external services. In these settings, auditability, permission control, rollback, and provenance are not optional. Modularity is therefore not only an engineering convenience; it is a requirement for safe and governable deployment, and an end-to-end policy still has to act through the same permission, verification, and audit surfaces we describe.

\paragraph{Objection 3: System-level evaluation is too expensive or environment-specific.}
System-level evaluation is indeed more expensive than static prompting benchmarks, and trace-level metrics are harder to standardize than endpoint accuracy. However, this is precisely why it is needed. Agents are deployed in environments where cost, latency, tool risk, memory drift, and verification overhead determine whether the system is usable. Evaluation protocols should expose these factors rather than abstract them away. The goal is not to replace simple benchmarks, but to complement them with measurements that reflect real agent operation.

\section{Conclusion}
\label{sec:conclusion}

Agentic AI is moving from isolated model inference to persistent system execution. As models are embedded into tools, memory stores, repositories, browsers, subagents, and external services, their behavior is increasingly shaped by the architecture around them. This paper has shown that future progress therefore requires \emph{system scaling}: improving how agents construct context, maintain trustworthy memory, route skills, verify actions, govern tools, communicate across roles, and evolve over time. Claude Code, OpenClaw, and CheetahClaws illustrate that comparable models projected onto different harnesses produce qualitatively different agents, and that the harness, not the model alone, is now a primary source of practical capability.

This does not diminish model scaling. Stronger foundation models remain essential, but model capability alone is no longer a sufficient unit of analysis for long-horizon agents. A mature science of agentic AI must study the full execution system: what it remembers, what it retrieves, what it exposes to the model, what actions it permits, what it verifies, and what it leaves auditable. Future benchmarks should therefore treat memory, context, skill routing across tools and subagents, orchestration, and verification-and-governance as first-class objects of design and evaluation, rather than measuring only one-shot success. Scaling the harness, alongside scaling the model, defines the next major bottleneck of agentic AI.

\section*{Acknowledgements}

We sincerely thank the open-source contributors to CheetahClaws for their valuable issue reports, pull requests, suggestions, and comments. We also thank Prof. Dawn Song and Prof. Costas Spanos for their great support.

\bibliography{reference}
\bibliographystyle{plain}


\appendix

\end{document}